\documentclass{article}

\usepackage[final]{bdu_2024}

\usepackage[utf8]{inputenc} % allow utf-8 input
\usepackage[T1]{fontenc}    % use 8-bit T1 fonts
\usepackage{hyperref}       % hyperlinks
\usepackage{url}            % simple URL typesetting
\usepackage{booktabs}       % professional-quality tables
\usepackage{amsfonts}       % blackboard math symbols
\usepackage{nicefrac}       % compact symbols for 1/2, etc.
\usepackage{microtype}      % microtypography
\usepackage{xcolor}         % colors
\usepackage{amsmath}
\usepackage{color, soul}
\usepackage{graphicx}
\usepackage{wrapfig}
\usepackage{multirow}
\usepackage{multicol}
\title{Cost-effective Reduced-Order Modeling via Bayesian Active Learning}

\author{%
  Amir Hossein Rahmati$^{1}$ \quad Nathan M. Urban$^{2}$ \quad Byung-Jun Yoon$^{1,2}$ \quad Xiaoning Qian$^{1,2}$\\\\
  $^1$ Texas A\&M University, College Station, TX \quad $^2$ Brookheaven National Laboratory, Upton, NY\\\\
  \texttt{\{amir\_hossein\_rahmati, bjyoon, xqian\}@tamu.edu}\\
  \texttt{\{nurban, byoon, xqian1\}@bnl.gov}\\
}

\begin{document}

\maketitle

\begin{abstract}
Machine Learning surrogates have been developed to accelerate solving systems dynamics of complex processes in different science and engineering applications. 
To faithfully capture %deliver a proper model of 
governing systems dynamics, these methods rely on large training datasets, % which is either unavailable or it brings about a heavy computational load, 
hence restricting their applicability in real-world problems. In this work, we propose BayPOD-AL, an active learning framework based on an uncertainty-aware Bayesian proper orthogonal decomposition (POD) approach, which aims to effectively learn reduced-order models from high-fidelity full-order models representing complex systems.
% uncertainty-quantification capabilities for reduced order models when modeling high-fidelity complex systems. 
% Any data selection strategy can be plugged into BayPOD-AL. In this work,
%Being flexible on choosing the data selection strategy and capable of utilizing uncertainty, we provide 
Experimental results 
% BayPOD-UAL and BayPOD-EAL, two strategies guided by uncertainty and error driven acquisition functions respectively. 
% estimated uncertainty and error respectively. 
%Results 
on predicting the temperature evolution over a rod demonstrate BayPOD-AL's effectiveness in suggesting the informative data and reducing computational cost related to constructing a training dataset compared to other uncertainty-guided active learning strategies. 
% We evaluate the efficiency the proposed method in our experiments. 
Furthermore, we demonstrate
%show the robustness of the proposed method's
BayPOD-AL's generalizability and efficiency by evaluating its performance on a dataset of higher temporal resolution than the training dataset.

% when the suggested data is selected based the settings of higher temporal resolution dataset than the training dataset. 

% on a dataset of higher temporal resolution than the training dataset.

% and the robustness of the proposed method in our experiments and using training and test datasets with different temporal resolution.

\end{abstract}

\section{Introduction}\label{Intro}

Many real-world decision-making problems benefit from proper modeling of high-dimensional complex systems dynamics. While traditional physics-principled computational models based on differential equations provide high-fidelity solutions, their prohibitive computational cost hinders their applications~\citep{10105938}. %\hl{shahin work}.  is a necessary requirement during the process of designing any such system
%, plus they do not have interpretable lower-dimensional embedding
With the abundance of data, from both high-fidelity simulations and real-world measurements, machine learning~(ML) surrogate models have become one of the exciting emerging solutions to learn the underlying governing dynamics of many real-world problems~\citep{10105938,verma2020surveymachinelearningapplied}. {These} surrogates have enabled efficient modeling of complex processes, instead of solely depending on solving their corresponding time-consuming, computationally expensive Ordinary or Partial Differential equation systems~(ODEs/PDEs). 
% With the abundance of data, from both high-fidelity simulations and real-world measurements, machine learning~(ML) surrogate models have become one of the exciting emerging solutions to efficiently learn the underlying governing dynamics of many real-world problems and model complex processes~\citep{10105938,verma2020surveymachinelearningapplied}, instead of solely depending on solving their corresponding time-consuming, computationally expensive Ordinary or Partial Differential equation systems~(ODEs/PDEs). 
More specifically, recent studies have been investigating the development of ML methods to accelerate these computations, in a spectrum from purely data-driven ML surrogates, physics-informed neural networks~(PINNs), to more recent hybrid models such as ML-augmented reduced-order models \citep{RAISSI2019686, SWISCHUK2019704, GUO2022115336, hirsh2021sparsifyingpriorsbayesianuncertainty, Lagaris_1998}. For instance, \citet{rudy2016datadrivendiscoverypartialdifferential} utilize sparse regression to learn a system's governing PDEs while \citet{raissi2017physicsinformeddeeplearning, raissi2018deephiddenphysicsmodels,Zhu_2018} consider black-box neural network models trained by ``physics-informed'' loss functions. 
Although these methods center their attention on providing accurate solutions, they often need retraining with a change of system settings, including parameters as well as initial and boundary conditions. {In \citet{DeGennaro_2019}, the authors proposed a two-step method to infer the model parameter posterior enabling uncertainty quantification after differential equation system identification.} %Different from BayPOD, it only infers the posterior for coefficients and it does not benefit from an end-to-end modeling.}  
A Bayesian framework for deriving reduced-order models (ROMs), BayPOD~\citep{10105938}, has been recently developed {in an end-to-end manner} based on proper orthogonal decomposition (POD), motivated by the idea of developing ML-augmented ROMs with embedded physics constraints~\citep{SWISCHUK2019704}. 
{ROM methods aim to derive physics-principled surrogates of high-fidelity complex models in significantly reduced lower-dimensional space to reduce the computational load while maintaining the desired accuracy}~\citep{HESTHAVEN201855}. By formulating ROM learning in a Bayesian framework, BayPOD is capable of quantifying the uncertainty in addition to providing approximate high-fidelity differential equation solutions. 
% in addition to providing the best approximates of the high-fidelity differential equation solutions, it is capable of quantifying the uncertainty.
This is a pivotal feature in scenarios where there is little observed data available, not atypical in science and engineering applications. 

%be accurate and interpretable load is still essential
Although these endeavors have facilitated new ML surrogates for more efficient modeling and forecasting of complex processes, they often require considerably large training datasets to achieve satisfactory performances. Acquiring such training data from high-fidelity full-order models~(FOMs) has heavy computational demands. To develop more data-efficient ML surrogates for ROM learning of complex systems, inspired by recent advancements of Active Learning (AL) in the ML community~\citep{6889457, ash2020deepbatchactivelearning, houlsby2011bayesianactivelearningclassification, wu2022entropybasedactivelearningobject, settles2009active, ren2021survey}, we here aim to bridge this gap by developing active learning for ROMs instead of constructing ROMs based on large batches of randomly generated FOM data. 
%wisely constructing the training dataset. reduce the training dataset size
More specifically, via iteratively suggesting the most informative data we intend to improve sample efficiency while preserving the underlying surrogate model's performance, reliability, and interoperability. 

%Throughout this study, by choosing due to its robustness and the design and use of ROMs with flexibility in , defining the acquisition function. Due to access to the prediction's an intuitive approach is Though it seems reasonable and it is easy to use, reducing the labeling cost depends on whether there is a model mismatch s and demonstrate the effectivenss
To develop and evaluate active learning for ROMs, we focus on BayPOD as the learned surrogate models by BayPOD come with their inherent uncertainty quantification capabilities in the adopted Bayesian learning framework. Here we promote a robust AL framework, BayPOD-AL, designed to showcase the feasibility and effectiveness of active learning for ROMs. With quantified uncertainty in BayPOD, we explore {different} uncertainty-based active learning strategies to utilize the estimated uncertainty for efficient guidance of the AL procedure. Recent studies in ML have suggested that uncertainty-based AL (UAL) methods can be unreliable in improving sample efficiency while optimizing the model's performance under specific scenarios~\citep{munjal2022robustreproducibleactivelearning,unknown, hacohen2022activelearningbudgetopposite, rahmati2024understandinguncertaintybasedactivelearning}. Exploring and evaluating different UAL strategies for ROM learning is critical to help understand and prevent potential performance degradation under the new ROM learning settings. In particular, as we focus on modeling complex systems behavior with ROMs, there is an inherent mismatch between the ROM surrogates and the systems to approximate. In \citet{rahmati2024understandinguncertaintybasedactivelearning}, the authors suggested error-based acquisition functions to alleviate the issues caused by the model mismatch. 
% targeting error as the acquisition function instead of the quantified uncertainty which can be inaccurate in representing the true objective function.
%When approximating the solutions of 
When developing ROMs for differential equation systems, the Mean Squared Error (MSE) is often considered as the target criterion. %Since the ground-truth solutions for unlabeled inputs are unavailable, error must be estimated. 
%Relying on literature and BayPOD-AL's capabilities, we present 
In this study, we propose and evaluate \textit{BayPOD-UAL} that is guided by an acquisition function depending solely on the estimated uncertainty and \textit{BayPOD-EAL} that relies on the estimated error. %for which its data selection strategy relies on the estimated error.  
% both uncertainty and error guided active learning strategies, BayPOD-UAL and BayPOD-EAL respectively. %BayPOD-UAL benefits from naturally uncertainty quantification capabilities of BayPOD while i
BayPOD-EAL, by taking advantage of the results in \citet{b713455c1b2c4ae28448b77823fe2a43}, is expected to achieve higher sample efficiency due to its objective-driven formulation directly targeting reducing MSE for ROM learning \citep{6473917,8574881, 9445102}. %\hl{cite our MOCU papers}. %the acquisition criterion depends on the estimated error. 
Throughout our experiments, we demonstrate the effectiveness of BayPOD-AL in improving sample efficiency when learning a BayPOD surrogate predicting the temperature evolution over a rod. Finally, by investigating its %BayPOD-AL's 
performance 
% when the most informative input is suggested using the setting of 
on a temporally high-resolution dataset, we further show its efficiency and robustness. 

% To further investigate the efficiency and robustness of BayPOD-AL, we evaluate its performance on \hl{?a dataset with higher temporal resolution than the training dataset.?}

\section{Active Learning for Reduced-Order Models}
Before delving into active learning for ROMs, we first briefly review ROMs, especially the family of POD-based ROMs, including BayPOD. %and AL in \ref{BG}.

% and BayPOD-EAL, two acquisition functions dependent on the estimated uncertainty and error respectively for ROMs. To formulate these functions, we briefly review AL and BayPOD in \ref{BG}.
%\subsection{Background}\label{BG}

\subsection{Reduced-Order Models}
%\hl{possible to write one short paragraph on that? moving from the BayPOD section? }
As mentioned in Section~\ref{Intro}, solving the full-order models (FOMs) to provide the high-fidelity solutions of the governing ODEs/PDEs is prohibitively expensive. Reduced-order models~(ROMs) are designed with the objective of reducing the computational cost by estimating FOM solutions in a lower-dimensional space, subject to keeping the information loss to a minimum~\citep{doi:10.1137/130932715,BESSELINK20134403,PENZL2006322,SWISCHUK2019704, Pant_2021}. 
Compared to pure data-driven ``black-box'' ML surrogates, learning ROMs of differential equation systems, naturally allows integrating the underlying scientific principles. We focus on proper orthogonal decomposition (POD), one of the most widely used model reduction methods to derive low-dimensional representations of the high-dimensional system states~\citep{SWISCHUK2019704}. 
Following the notations from \citet{10105938}, consider a function $f: \mathcal{X}\times\mathcal{T}\times\mathcal{P} \rightarrow \mathbb{R}$, where $\mathcal{X}$, the spatial domain, $\mathcal{T}$, the time domain, and $\mathcal{P}$, the input domain are mapped to a physical field. Denoting a snapshot $\boldsymbol{f}(t;\mathbf{p})\in \mathbb{R}^{n_x}$ as the discretized spatial domain at time $t$ and input model parameters $\mathbf{p}$, with $n_t$ different time points and $n_\mathbf{p}$ different inputs, the POD bases can be obtained via singular value decomposition (SVD). Specifically, defining $F = [\boldsymbol{f}(t_i;\mathbf{p}_j)]\in \mathbb{R}^{n_s\times n_x}$ with total $n_s = n_tn_p$ snapshots, the SVD is $F=V\Sigma W$, which enables approximating field $f$ by the $K$-dimension POD basis for any input model parameters.

\subsection{Bayesian POD~(BayPOD)}\label{BPOD-S2}

%In this section, we borrow 
%the notations from \citet{10105938}. Consider a function $f: \mathcal{X}\times\mathcal{T}\times\mathcal{P} \rightarrow \mathbb{R}$, where 

BayPOD focuses on simultaneously deriving low-dimensional projection and mapping from input parameters of the full-order model to latent projection coefficients. To learn the map $\boldsymbol{\alpha}:\mathcal{P}\times\mathcal{T}\rightarrow\mathcal{A}$, 
% ML models have been employed  where $\boldsymbol{\alpha}(\mathbf{p},t)\in \mathbb{R}^K$, and $\mathbf{p}\in \mathbb{R}^m$ are $m$-dimensional input parameters. In this study, 
we consider neural network mappings.  
To account for physics constraints, BayPOD reformulates the linear representation of each snapshot's approximation, $\Tilde{\boldsymbol{f}}$, constituting of the POD approximation and a \emph{particular solution} given the corresponding initial and boundary conditions $\Tilde{\boldsymbol{f}}(t;\mathbf{p}) = \Bar{\boldsymbol{f}} + \sum_{k=1}^{K} \Bar{\boldsymbol{v}}_k{\alpha}_k(t;\mathbf{p})$, 
\iffalse
\begin{equation}
    \Tilde{\boldsymbol{f}}(t;\mathbf{p}) = \Bar{\boldsymbol{f}} + \sum_{k=1}^{K} \Bar{\boldsymbol{v}}_k{\alpha}_k(t;\mathbf{p}), \label{eq:approxpod}
\end{equation}
\fi
where $\alpha_k(t;\mathbf{p})$ is the corresponding \emph{learnable} POD expansion coefficient, and $\Bar{\boldsymbol{v}}_k \in \mathbb{R}^{n_x}$ is the $k$-th left singular vector of $F$ as the POD bases that satisfy homogeneous boundary conditions. Here $\Bar{\boldsymbol{f}}$ is the \emph{particular solution} given the set of potentially inhomogeneous initial and boundary conditions for better embedding physics constraints~\citep{SWISCHUK2019704,10105938}. % that can model  % and $\Bar{\boldsymbol{v}}$  are the POD bases that satisfy homogeneous boundary conditions. 
%models dynamics with homogeneous boundary conditions while part is modeled. Denoting the field is modeled
The corresponding field value of snapshot $s$ at the spatial point $x$, $\Tilde{f}_{sx}$, is modeled as a normal random variable: 
\begin{equation}
    \Tilde{f}_{sx} \sim \mathbf{N}(\boldsymbol{u}_x^{\top}\boldsymbol{\alpha}_s, \gamma_x^{-1})
\end{equation}
with $\boldsymbol{u}_x\in \mathbb{R}^K$ the $K$-dimensional POD basis at position $x$, and $\gamma_x^{-1}$ the variance at $x$. {Using mean-field variational inference, BayPOD finds the variational posterior of model parameters. Due to its generative nature, it provides uncertainty estimates of predicted system dynamics in different setups, which is the enabler of optimal and adaptive decision making, active learning for sample} 
\begin{wrapfigure}{r}{0.4\textwidth}\vspace{-3mm}
    \centering
\includegraphics[width=0.4\textwidth]{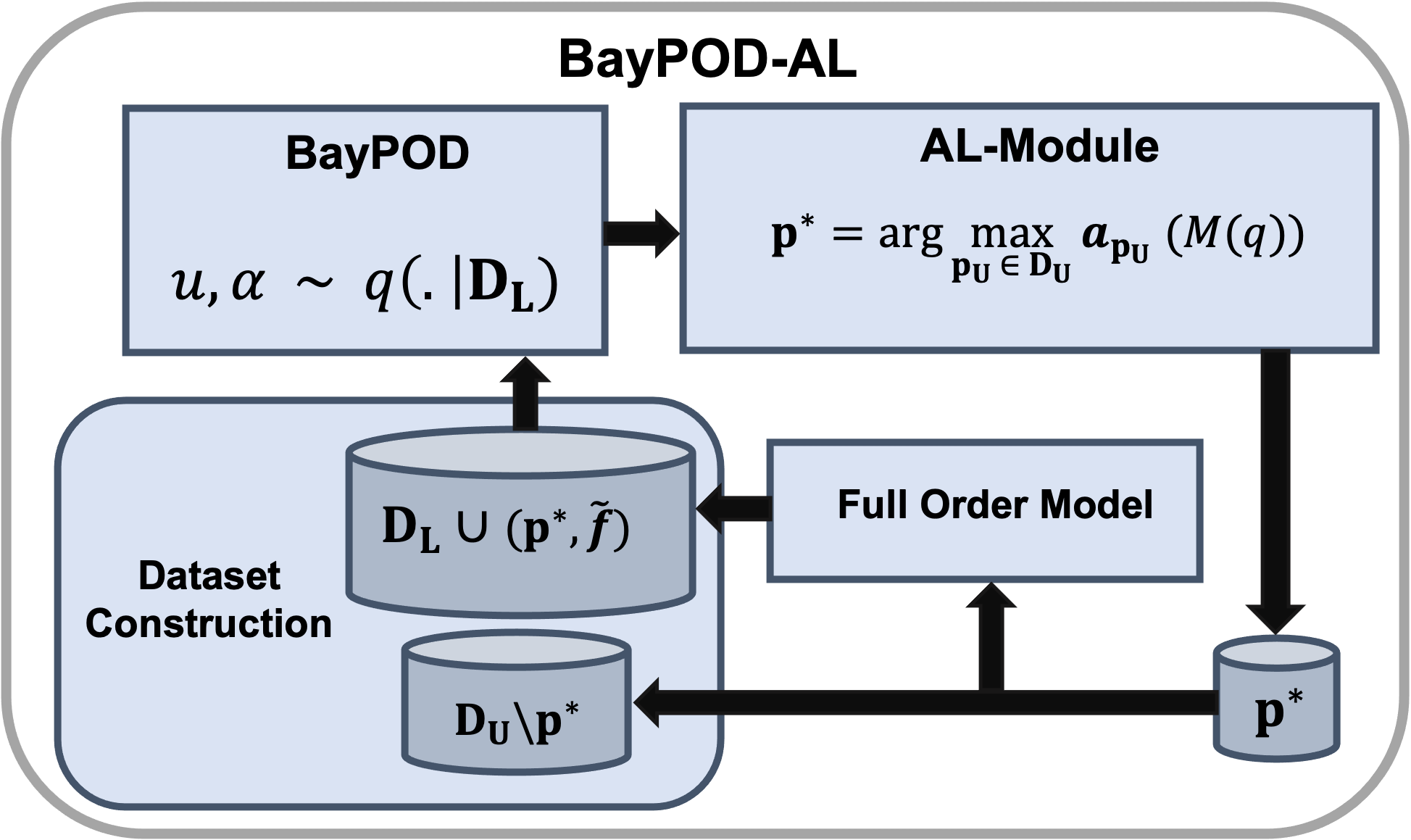}\vspace{-2mm}
    \caption{A schematic illustration of the proposed BayPOD-AL framework.}
    \label{fig:bpod-al-diag}\vspace{-12mm}
\end{wrapfigure}
efficiency in this work.
% By defining $\gamma_\alpha$ as $\boldsymbol{\alpha}_s$'s precision parameter,  
% using variational inference, and assuming independence between variational parameters, $q(\boldsymbol{u}, \boldsymbol{\alpha}, \boldsymbol{\gamma}) = q(\boldsymbol{u})q(\boldsymbol{\alpha})q(\boldsymbol{\gamma})$ with  variational distribution $q(\cdot)$, the variational posteriors 
% % of POD basis vectors, coefficients, and precision parameters 
% are inferred. Due to its generative nature, BayPOD provides uncertainty estimates of predicted systems dynamics in different setups, which is the enabler of optimal and adaptive decision making, active learning for sample efficiency in this work. 

\subsection{BayPOD-AL}\label{BPOD-AL-S2}

We now present our active learning framework, BayPOD-AL, by exploring different uncertainty-based active learning~(UAL) strategies, leveraging the inherent UQ capabilities of BayPOD. 
%\subsubsection{Active Learning}
Consider $\mathbf{D_L}$ and $\mathbf{D_U}$ the iteratively updated `labeled' and `unlabeled' datasets corresponding to collected snapshots from high-fidelity FOMs and the FOM settings without actual simulated snapshots, respectively. BayPOD-AL focuses on reducing the cost of running FOM simulations to train BayPOD by choosing the most informative FOM settings in $\mathbf{D_U}$ to collect new BayPOD training data from FOM based on an acquisition function $\boldsymbol{a}$. Specifically, BayPOD-AL iteratively queries for snapshots, solutions to corresponding FOM differential equations, with the considered most informative settings, the corresponding input FOM parameters in this work $\mathbf{p}^* = \arg \max_{\mathbf{p_U} \in \mathbf{D_U}} \boldsymbol{a}(q(\cdot|\mathbf{D_L}), \mathbf{p_U})$. Here the acquisition function $\boldsymbol{a}(q(\cdot|\mathbf{D_L}), \mathbf{p_U})$ guides the AL procedure considering the potential uncertainty of iteratively updated BayPOD models. {At each AL iteration, with the currently learned BayPOD model, BayPOD-AL determines the most informative input FOM parameter in $\mathbf{D_U}$, i.e. $\mathbf{p}^*$, as the output to query additional training snapshots from the FOM.} Note that BayPOD directly provides the model posterior given the labeled data or simulated snapshots, $q(\cdot|\mathbf{D_L})$. Until the trained BayPOD model reaches a satisfactory performance, BayPOD-AL continues to add new simulated FOM snapshots to $\mathbf{D_L}$. 
%In this study, for UAL we consider the estimated variance of BayPOD's approximated snapshots as the uncertainty measure for acquisition function. 

In our BayPOD-AL framework, {due to solving a homogeneous problem that frees us from worrying about boundary/initial constraints, we only} search for the most `informative' input FOM parameters $\mathbf{p}^* \in \mathbf{D_U} \subset \mathcal{P}$. Assuming that $\mathbf{D_U}$ contains $n_\mathbf{p}$ inputs, and for each input $\mathbf{p_U} \in \mathbf{D_U}$, we want to query FOM solutions for $n_t^{\mathbf{p}}$ \textbf{fixed} time points. At each AL iteration, we compute the acquisition function for a batch of $n_t^{\mathbf{p}}$ snapshots for \textbf{each} input. Finally, the most informative snapshots from FOMs will be appended to $\mathbf{D_L}$ for the next AL and model update iteration. %of the input will be acquired 
This process continues until the model reaches the desired performance. 
Having access to the BayPOD's variational posterior $q(\cdot|\mathbf{D_L})$, we define a measure function $M^{(t)}(q(\cdot|\mathbf{D_L},\mathbf{p_U},x)): \mathcal{U}\times\mathcal{A}\rightarrow\mathbb{R}$ estimating the `informativeness' of snapshots at input $\mathbf{p_U}$, time $t$, and position $x$. By taking the average over all snapshots given $\mathbf{p_U}$, we define the following acquisition function evaluating such `informativeness' based on the adopted measure function: % for each $\mathbf{p_U}$: %each unlabeled input's significance is calculated as, 
\begin{equation}\label{bpod-al}
 \boldsymbol{a}_\mathbf{p_U}^{(i)} = \frac{1}{n_t^{\mathbf{p}}} \sum_t^{n_t^{\mathbf{p}}} \frac{1}{n_x}\sum_x^{n_x} M^{(i,t)}(q(\cdot|\mathbf{D_L},\mathbf{p_U},x)), \qquad i \in \{0, \dots, n_\mathbf{p}\}
\end{equation}
BayPOD-AL then queries the corresponding FOM for snapshots by the most `informative' input: 
%significant input is then selected as follows, 
\begin{equation}\label{p-star}
    \mathbf{p}^{*} = \mathbf{p_U}^{(i)} = \arg\max_{i} \boldsymbol{a}_\mathbf{p_U}^{(i)}. 
\end{equation}
Figure~\ref{fig:bpod-al-diag} provides a schematic illustration of BayPOD-AL. %provides a schematic representation of this framework.
As discussed in \citet{hino2020activelearningproblemsettings, pmlr-v130-zhao21c, rahmati2024understandinguncertaintybasedactivelearning}, the acquisition function plays a critical role in achieving desired AL efficiency. To derive efficient active learning for BayPOD, we evaluate two different choices by considering:  %consider two general approaches in designing such functions: 
%1) acquisition functions driven entirely based on 
1) \emph{BayPOD-UAL}: the predictive model uncertainty based on the estimated posterior predictive variance from BayPOD; and 2) \emph{BayPOD-EAL}: the estimated approximation error, directly targeting the ROM learning objective, for which we estimate the upper bound of the approximation error and utilize that as the measure function in~\eqref{bpod-al}. %{which we denote as BayPOD-UAL and BayPOD-EAL respectively}. {In this study, to perform BayPOD-UAL, we consider the estimated variance as the measure function. For implementing BayPOD-EAL, we focus on estimating the upper bound of the error and utilize that as the measure function. 
Details of each strategy are provided in Appendix~\ref{UAL} and~\ref{EAL}.   

\section{Experiments}\label{exps}
Using the same example as in \citet{10105938}, we implement our BayPOD-AL on predicting the evolution of temperature fields, $f$, {with heat diffusivity parameter, $\kappa$,} over a rod of length $L$ with time-dependent boundary conditions. 
{In the following}, we report the performance statistics of five runs of experiments for each of AL algorithms. Each run is different only in the initial $\mathbf{D_L}$. We compare the performance of BayPOD-UAL and BayPOD-EAL with the random sampling strategy which selects the next $\kappa$ for FOM simulations randomly at each iteration. %round of data selection.

\begin{wrapfigure}{r}{0.5\textwidth}
    \centering\vspace{-5mm}
    \includegraphics[width=0.5\textwidth]{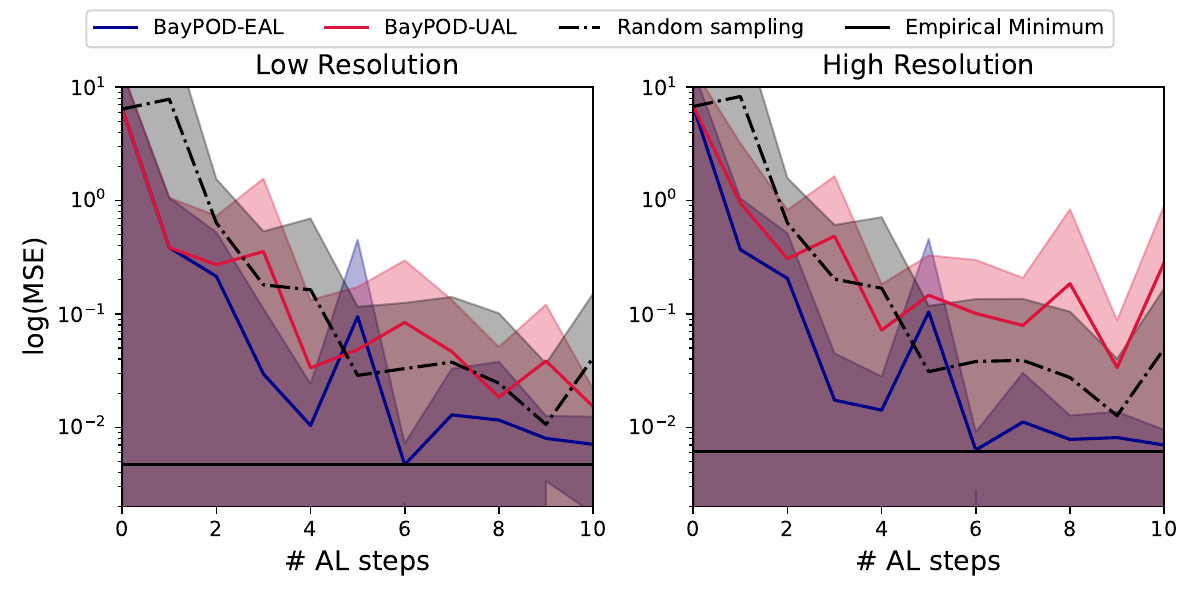}\vspace{-5mm}
    \caption{Performance comparison of BayPOD-EAL, BayPOD-UAL, and random sampling strategies on `low' (left) and `high' (right) temporal resolution test datasets. 
    % \hl{Need to be changed} 
    }\vspace{-3mm}
    %\hl{two plots are the same? please also change the text in plots into BayPOD-UAL, etc? }}
    \label{fig:erAL}
\end{wrapfigure}

% As mentioned in Section~\ref{setting}, 
We train BayPOD on the temporally low-resolution training snapshots and report the AL performances on both temporally low-($n_t^\mathbf{p} = 50$) and high-resolution ($n_t^\mathbf{p} = 200$) test datasets. Figure \ref{fig:erAL} compares the performance of BayPOD-EAL, BayPOD-UAL, and random sampling strategies. {Table}~\ref{table1}{ summarizes the performance statistics of our results after 5 AL iterations.} It is clear after 5 AL iterations (250 new snapshots), BayPOD-EAL leads to a model with the best empirical performance. Over the first 10 AL iterations, on average it has $4.7\times$ and $9\times$ lower MSE than BayPOD-UAL, and $6\times$ and $6.5\times$ lower MSE than random sampling when evaluated on low- and high-resolution datasets respectively. This further demonstrates the robustness of BayPOD-EAL, in leading the model to optimal performance even when the training data resolution differs from the test data, owing to it being objective-driven. 
As mentioned in Section~\ref{Intro}, due to the inherent model mismatch, uncertainty alone cannot efficiently reduce the labeling cost for learning ROMs. %Similar to observations in the regression case study\hl{myself?}, 
On both datasets, BayPOD-UAL grants an initial performance lead compared to random sampling, but it soon slows down with comparable performance for the low-resolution dataset and worse performance for the high-resolution dataset. %It is worth noting that 
After 15 to 20 AL iterations, both BayPOD-EAL and random sampling provide comparable performance, meaning that BayPOD-AL reduces the computational cost related to training data by a factor of $3$ to $4$, demonstrating its cost-effectiveness %of BayPOD-AL 
in learning ROMs.%\vspace{-2mm}

\begin{table}[]
    \centering
    \caption{Mean and standard deviation (std) of approximation error at the $6^{th}$ AL iterations from five random runs of experiments. Best results are in bold.}
    \begin{tabular}{|c||c|c||c|c|}\hline
         \multirow{2}{*}{Method}&\multicolumn{2}{|c||}{Low Resolution}&\multicolumn{2}{|c|}{High Resolution}  \\ \cline{2-5}
         &mean & std& mean & std \\ \hline\hline
         BayPOD-EAL&$\mathbf{4.68 \times 10^{-3}}$&$\mathbf{1.3 \times 10^{-3}}$&$\mathbf{6.3 \times 10^{-3}}$& $\mathbf{1.79\times 10^{-3}}$\\ \hline
         BayPOD-UAL&$8.42 \times 10^{-2}$&$1.08 \times 10^{-1}$&$1.01 \times 10^{-1}$&$1.01 \times 10^{-1}$ \\ \hline
         Random Sampling&$3.29\times 10^{-2}$&$4.7 \times 10^{-2}$&$3.8 \times 10^{-2}$&$4.95 \times 10^{-2}$ \\ \hline
    \end{tabular}
    \label{table1}
\end{table}

\section{Conclusion}%\vspace{-2mm}

To model the complex systems dynamics, ML surrogate models have shown promising performance. However, the prohibitive cost of acquiring the solutions of the high-fidelity FOMs representing them 
% ground-truth solutions for these high-dimensional complex systems 
hinders the applicability of these models in real-world scenarios. Inspired by recent advances in AL, we present the BayPOD-AL framework, which iteratively suggests the most informative data that can boost the surrogate model's performance. The Bayesian framework explicitly updates the model posterior, with which different objective-driven acquisition functions targeting ROM learning can be adopted to achieve efficient AL. %\hl{Accessing to the model's posteriors, it provides ample opportunities in defining acquisition functions, facilitating its use. (Note: Rephrase to make the meaning clearer)} 
Our experiments demonstrate the robustness of the proposed framework as well as its efficacy in reducing the cost of learning ROMs, thereby improving their applicability in real-world problems.
%instead of solving the governing equation for any arbitrary inputs which can be extremely expensive and most likely result in redundant information.
%The efficacy of the proposed framework on high-resolution data demonstrates its robustness, reliability, and applicability in potential real-world problems.

\noindent\textbf{Acknowledgements} This work was supported in part by the U.S. National Science Foundation~(NSF) grants IIS-2212419; and by the U.S. Department of Engergy~(DOE) Office of Science, Advanced Scientific Computing Research (ASCR) M2DT Mathematical Multifaceted Integrated Capability Center~(MMICC) under Award B\&R\# KJ0401010/FWP\# CC130, program manager W. Spotz, and Award DE-SC0012704, program manager Margaret Lentz. Portions of this research were conducted with the advanced computing resources provided by Texas A\&M High Performance Research Computing.

\bibliography{ref}

\begin{thebibliography}{33}
\providecommand{\natexlab}[1]{#1}
\providecommand{\url}[1]{\texttt{#1}}
\expandafter\ifx\csname urlstyle\endcsname\relax
  \providecommand{\doi}[1]{doi: #1}\else
  \providecommand{\doi}{doi: \begingroup \urlstyle{rm}\Url}\fi

\bibitem[Ash et~al.(2020)Ash, Zhang, Krishnamurthy, Langford, and
  Agarwal]{ash2020deepbatchactivelearning}
J.~T. Ash, C.~Zhang, A.~Krishnamurthy, J.~Langford, and A.~Agarwal.
\newblock Deep batch active learning by diverse, uncertain gradient lower
  bounds, 2020.
\newblock URL \url{https://arxiv.org/abs/1906.03671}.

\bibitem[Benner et~al.(2015)Benner, Gugercin, and
  Willcox]{doi:10.1137/130932715}
P.~Benner, S.~Gugercin, and K.~Willcox.
\newblock A survey of projection-based model reduction methods for parametric
  dynamical systems.
\newblock \emph{SIAM Review}, 57\penalty0 (4):\penalty0 483--531, 2015.
\newblock \doi{10.1137/130932715}.
\newblock URL \url{https://doi.org/10.1137/130932715}.

\bibitem[Besselink et~al.(2013)Besselink, Tabak, Lutowska, {van de Wouw},
  Nijmeijer, Rixen, Hochstenbach, and Schilders]{BESSELINK20134403}
B.~Besselink, U.~Tabak, A.~Lutowska, N.~{van de Wouw}, H.~Nijmeijer, D.~Rixen,
  M.~Hochstenbach, and W.~Schilders.
\newblock A comparison of model reduction techniques from structural dynamics,
  numerical mathematics and systems and control.
\newblock \emph{Journal of Sound and Vibration}, 332\penalty0 (19):\penalty0
  4403--4422, 2013.
\newblock ISSN 0022-460X.
\newblock \doi{https://doi.org/10.1016/j.jsv.2013.03.025}.
\newblock URL
  \url{https://www.sciencedirect.com/science/article/pii/S0022460X1300285X}.

\bibitem[Boluki et~al.(2019)Boluki, Qian, and Dougherty]{8574881}
S.~Boluki, X.~Qian, and E.~R. Dougherty.
\newblock Experimental design via generalized mean objective cost of
  uncertainty.
\newblock \emph{IEEE Access}, 7:\penalty0 2223--2230, 2019.
\newblock \doi{10.1109/ACCESS.2018.2886576}.

\bibitem[Boluki et~al.(2024)Boluki, Dadaneh, Dougherty, and Qian]{10105938}
S.~Boluki, S.~Z. Dadaneh, E.~R. Dougherty, and X.~Qian.
\newblock Bayesian proper orthogonal decomposition for learnable reduced-order
  models with uncertainty quantification.
\newblock \emph{IEEE Transactions on Artificial Intelligence}, 5\penalty0
  (3):\penalty0 1162--1173, 2024.
\newblock \doi{10.1109/TAI.2023.3268609}.

\bibitem[DeGennaro et~al.(2019)DeGennaro, Urban, Nadiga, and
  Haut]{DeGennaro_2019}
A.~M. DeGennaro, N.~M. Urban, B.~T. Nadiga, and T.~Haut.
\newblock Model structural inference using local dynamic operators.
\newblock \emph{International Journal for Uncertainty Quantification},
  9\penalty0 (1):\penalty0 59–83, 2019.
\newblock ISSN 2152-5080.
\newblock \doi{10.1615/int.j.uncertaintyquantification.2019025828}.
\newblock URL
  \url{http://dx.doi.org/10.1615/Int.J.UncertaintyQuantification.2019025828}.

\bibitem[Guo et~al.(2022)Guo, McQuarrie, and Willcox]{GUO2022115336}
M.~Guo, S.~A. McQuarrie, and K.~E. Willcox.
\newblock Bayesian operator inference for data-driven reduced-order modeling.
\newblock \emph{Computer Methods in Applied Mechanics and Engineering},
  402:\penalty0 115336, 2022.
\newblock ISSN 0045-7825.
\newblock \doi{https://doi.org/10.1016/j.cma.2022.115336}.
\newblock URL
  \url{https://www.sciencedirect.com/science/article/pii/S0045782522004273}.
\newblock A Special Issue in Honor of the Lifetime Achievements of J. Tinsley
  Oden.

\bibitem[Hacohen et~al.(2022)Hacohen, Dekel, and
  Weinshall]{hacohen2022activelearningbudgetopposite}
G.~Hacohen, A.~Dekel, and D.~Weinshall.
\newblock Active learning on a budget: Opposite strategies suit high and low
  budgets, 2022.
\newblock URL \url{https://arxiv.org/abs/2202.02794}.

\bibitem[Hesthaven and Ubbiali(2018)]{HESTHAVEN201855}
J.~Hesthaven and S.~Ubbiali.
\newblock Non-intrusive reduced order modeling of nonlinear problems using
  neural networks.
\newblock \emph{Journal of Computational Physics}, 363:\penalty0 55--78, 2018.
\newblock ISSN 0021-9991.
\newblock \doi{https://doi.org/10.1016/j.jcp.2018.02.037}.
\newblock URL
  \url{https://www.sciencedirect.com/science/article/pii/S0021999118301190}.

\bibitem[Hino(2020)]{hino2020activelearningproblemsettings}
H.~Hino.
\newblock Active learning: Problem settings and recent developments, 2020.
\newblock URL \url{https://arxiv.org/abs/2012.04225}.

\bibitem[Hirsh et~al.(2021)Hirsh, Barajas-Solano, and
  Kutz]{hirsh2021sparsifyingpriorsbayesianuncertainty}
S.~M. Hirsh, D.~A. Barajas-Solano, and J.~N. Kutz.
\newblock Sparsifying priors for bayesian uncertainty quantification in model
  discovery, 2021.
\newblock URL \url{https://arxiv.org/abs/2107.02107}.

\bibitem[Houlsby et~al.(2011)Houlsby, Huszár, Ghahramani, and
  Lengyel]{houlsby2011bayesianactivelearningclassification}
N.~Houlsby, F.~Huszár, Z.~Ghahramani, and M.~Lengyel.
\newblock Bayesian active learning for classification and preference learning,
  2011.
\newblock URL \url{https://arxiv.org/abs/1112.5745}.

\bibitem[Lagaris et~al.(1998)Lagaris, Likas, and Fotiadis]{Lagaris_1998}
I.~Lagaris, A.~Likas, and D.~Fotiadis.
\newblock Artificial neural networks for solving ordinary and partial
  differential equations.
\newblock \emph{IEEE Transactions on Neural Networks}, 9\penalty0 (5):\penalty0
  987–1000, 1998.
\newblock ISSN 1045-9227.
\newblock \doi{10.1109/72.712178}.
\newblock URL \url{http://dx.doi.org/10.1109/72.712178}.

\bibitem[Munjal et~al.(2022)Munjal, Hayat, Hayat, Sourati, and
  Khan]{munjal2022robustreproducibleactivelearning}
P.~Munjal, N.~Hayat, M.~Hayat, J.~Sourati, and S.~Khan.
\newblock Towards robust and reproducible active learning using neural
  networks, 2022.
\newblock URL \url{https://arxiv.org/abs/2002.09564}.

\bibitem[Pant et~al.(2021)Pant, Doshi, Bahl, and Barati~Farimani]{Pant_2021}
P.~Pant, R.~Doshi, P.~Bahl, and A.~Barati~Farimani.
\newblock Deep learning for reduced order modelling and efficient temporal
  evolution of fluid simulations.
\newblock \emph{Physics of Fluids}, 33\penalty0 (10), Oct. 2021.
\newblock ISSN 1089-7666.
\newblock \doi{10.1063/5.0062546}.
\newblock URL \url{http://dx.doi.org/10.1063/5.0062546}.

\bibitem[Penzl(2006)]{PENZL2006322}
T.~Penzl.
\newblock Algorithms for model reduction of large dynamical systems.
\newblock \emph{Linear Algebra and its Applications}, 415\penalty0
  (2):\penalty0 322--343, 2006.
\newblock ISSN 0024-3795.
\newblock \doi{https://doi.org/10.1016/j.laa.2006.01.007}.
\newblock URL
  \url{https://www.sciencedirect.com/science/article/pii/S0024379506000371}.
\newblock Special Issue on Order Reduction of Large-Scale Systems.

\bibitem[Rahmati et~al.(2024)Rahmati, Fan, Zhou, Urban, Yoon, and
  Qian]{rahmati2024understandinguncertaintybasedactivelearning}
A.~H. Rahmati, M.~Fan, R.~Zhou, N.~M. Urban, B.-J. Yoon, and X.~Qian.
\newblock Understanding uncertainty-based active learning under model mismatch,
  2024.
\newblock URL \url{https://arxiv.org/abs/2408.13690}.

\bibitem[Raissi(2018)]{raissi2018deephiddenphysicsmodels}
M.~Raissi.
\newblock Deep hidden physics models: Deep learning of nonlinear partial
  differential equations, 2018.
\newblock URL \url{https://arxiv.org/abs/1801.06637}.

\bibitem[Raissi et~al.(2017)Raissi, Perdikaris, and
  Karniadakis]{raissi2017physicsinformeddeeplearning}
M.~Raissi, P.~Perdikaris, and G.~E. Karniadakis.
\newblock Physics informed deep learning (part i): Data-driven solutions of
  nonlinear partial differential equations, 2017.
\newblock URL \url{https://arxiv.org/abs/1711.10561}.

\bibitem[Raissi et~al.(2019)Raissi, Perdikaris, and Karniadakis]{RAISSI2019686}
M.~Raissi, P.~Perdikaris, and G.~Karniadakis.
\newblock Physics-informed neural networks: A deep learning framework for
  solving forward and inverse problems involving nonlinear partial differential
  equations.
\newblock \emph{Journal of Computational Physics}, 378:\penalty0 686--707,
  2019.
\newblock ISSN 0021-9991.
\newblock \doi{https://doi.org/10.1016/j.jcp.2018.10.045}.
\newblock URL
  \url{https://www.sciencedirect.com/science/article/pii/S0021999118307125}.

\bibitem[Ren et~al.(2021)Ren, Xiao, Chang, Huang, Li, Gupta, Chen, and
  Wang]{ren2021survey}
P.~Ren, Y.~Xiao, X.~Chang, P.-Y. Huang, Z.~Li, B.~B. Gupta, X.~Chen, and
  X.~Wang.
\newblock A survey of deep active learning, 2021.

\bibitem[Rudy et~al.(2016)Rudy, Brunton, Proctor, and
  Kutz]{rudy2016datadrivendiscoverypartialdifferential}
S.~H. Rudy, S.~L. Brunton, J.~L. Proctor, and J.~N. Kutz.
\newblock Data-driven discovery of partial differential equations, 2016.
\newblock URL \url{https://arxiv.org/abs/1609.06401}.

\bibitem[Saifullah et~al.(2022)Saifullah, Agne, Dengel, and Ahmed]{unknown}
S.~Saifullah, S.~Agne, A.~Dengel, and S.~Ahmed.
\newblock Analyzing the potential of active learning for document image
  classification, 11 2022.

\bibitem[Savvides et~al.(2024)Savvides, Luu, and
  Puolam{\"a}ki]{b713455c1b2c4ae28448b77823fe2a43}
R.~Savvides, H.~Luu, and K.~Puolam{\"a}ki.
\newblock Error bounds for any regression model using {G}aussian processes with
  gradient information.
\newblock \emph{Proceedings of Machine Learning Research}, 238:\penalty0
  397--405, 2024.
\newblock ISSN 2640-3498.
\newblock URL \url{http://aistats.org/aistats2024/}.
\newblock International Conference on Artificial Intelligence and Statistics,
  AISTATS ; Conference date: 02-05-2024 Through 04-05-2024.

\bibitem[Settles(2009)]{settles2009active}
B.~Settles.
\newblock Active learning literature survey.
\newblock 2009.

\bibitem[Swischuk et~al.(2019)Swischuk, Mainini, Peherstorfer, and
  Willcox]{SWISCHUK2019704}
R.~Swischuk, L.~Mainini, B.~Peherstorfer, and K.~Willcox.
\newblock Projection-based model reduction: Formulations for physics-based
  machine learning.
\newblock \emph{Computers \& Fluids}, 179:\penalty0 704--717, 2019.
\newblock ISSN 0045-7930.
\newblock \doi{https://doi.org/10.1016/j.compfluid.2018.07.021}.
\newblock URL
  \url{https://www.sciencedirect.com/science/article/pii/S0045793018304250}.

\bibitem[Verma(2020)]{verma2020surveymachinelearningapplied}
S.~Verma.
\newblock A survey on machine learning applied to dynamic physical systems,
  2020.
\newblock URL \url{https://arxiv.org/abs/2009.09719}.

\bibitem[Wang and Shang(2014)]{6889457}
D.~Wang and Y.~Shang.
\newblock A new active labeling method for deep learning.
\newblock In \emph{2014 International Joint Conference on Neural Networks
  (IJCNN)}, pages 112--119, 2014.
\newblock \doi{10.1109/IJCNN.2014.6889457}.

\bibitem[Wu et~al.(2022)Wu, Chen, and
  Huang]{wu2022entropybasedactivelearningobject}
J.~Wu, J.~Chen, and D.~Huang.
\newblock Entropy-based active learning for object detection with progressive
  diversity constraint, 2022.
\newblock URL \url{https://arxiv.org/abs/2204.07965}.

\bibitem[Yoon et~al.(2013)Yoon, Qian, and Dougherty]{6473917}
B.-J. Yoon, X.~Qian, and E.~R. Dougherty.
\newblock Quantifying the objective cost of uncertainty in complex dynamical
  systems.
\newblock \emph{IEEE Transactions on Signal Processing}, 61\penalty0
  (9):\penalty0 2256--2266, 2013.
\newblock \doi{10.1109/TSP.2013.2251336}.

\bibitem[Yoon et~al.(2021)Yoon, Qian, and Dougherty]{9445102}
B.-J. Yoon, X.~Qian, and E.~R. Dougherty.
\newblock Quantifying the multi-objective cost of uncertainty.
\newblock \emph{IEEE Access}, 9:\penalty0 80351--80359, 2021.
\newblock \doi{10.1109/ACCESS.2021.3085486}.

\bibitem[Zhao et~al.(2021)Zhao, Dougherty, Yoon, J.~Alexander, and
  Qian]{pmlr-v130-zhao21c}
G.~Zhao, E.~Dougherty, B.-J. Yoon, F.~J.~Alexander, and X.~Qian.
\newblock Bayesian active learning by soft mean objective cost of uncertainty.
\newblock In A.~Banerjee and K.~Fukumizu, editors, \emph{Proceedings of The
  24th International Conference on Artificial Intelligence and Statistics},
  volume 130 of \emph{Proceedings of Machine Learning Research}, pages
  3970--3978. PMLR, 13--15 Apr 2021.
\newblock URL \url{https://proceedings.mlr.press/v130/zhao21c.html}.

\bibitem[Zhu and Zabaras(2018)]{Zhu_2018}
Y.~Zhu and N.~Zabaras.
\newblock Bayesian deep convolutional encoder–decoder networks for surrogate
  modeling and uncertainty quantification.
\newblock \emph{Journal of Computational Physics}, 366:\penalty0 415–447,
  Aug. 2018.
\newblock ISSN 0021-9991.
\newblock \doi{10.1016/j.jcp.2018.04.018}.
\newblock URL \url{http://dx.doi.org/10.1016/j.jcp.2018.04.018}.

\end{thebibliography}
% \bibliographystyle{nature}

%%%%%%%%%%%%%%%%%%%%%%%%%%%%%%%%%%%%%%%%%%%%%%%%%%%%%%%%%%%%

\appendix

\section{BayPOD: Posterior Inference}
As mentioned in Section \ref{BPOD-S2} the response is modeled as a normally-distributed random variable, $\Tilde{f}_{sx} \sim \mathbf{N}(\boldsymbol{u}^\top_x\boldsymbol{\alpha}_s, \gamma^{-1}_x)$, with $\boldsymbol{\alpha_s}, \boldsymbol{u}_x \in \mathbb{R}^K$ the $K$-dimensional POD basis vector at $x$, and $K$ POD coefficients for snapshot $s$. Considering zero-mean normal priors on POD basis and coefficients, i.e., $\boldsymbol{u}_x \sim \mathbf{N}(0,I)$ and $\boldsymbol{\alpha}_s \sim \mathbf{N}(0, \gamma_\alpha^{-1}I)$, as well as conjugate gamma distributions over the precision parameters, $\gamma_\alpha, \gamma_x \sim \text{Gamma}(1,1)$, the model is completed. 

Using variational inference, the posteriors over parameters are inferred. More specifically, variational distributions $q(\cdot)$ is put over model parameters with the independence assumption, that is $q(\boldsymbol{u},\boldsymbol{\alpha}, \boldsymbol{\gamma}) = q(\boldsymbol{u})q(\boldsymbol{\alpha})q(\boldsymbol{\gamma})$. Since we consider using a neural network (NN) for coefficient mapping, the variational distribution over $\boldsymbol{\alpha}_s$ can be defined as $q(\boldsymbol{\alpha}_s) = \mathbf{N}(\boldsymbol{\alpha}_s;\mu_w(\mathbf{p}), \Sigma_w(\mathbf{p}))$, with $\mu_w$ and $\Sigma_w$ the mean and covariance matrix of NN form with weights $w$. {Similar as in} \citet{10105938}{, we employ the same NN architecture with two hidden layers, 50 nodes per layer, and using rectified linear unit (ReLU) activation functions.} To benefit from conjugate priors, the variational posteriors for POD basis and precision parameters are set to be normal and gamma distributions. Finally, by minimizing the Kullback-Leibler (KL) divergence between the variational posteriors and the true posteriors, the optimal parameters are obtained. 

\section{BayPOD-UAL: Uncertainty-guided AL}\label{UAL}
The acquisition function can be utilized by adopting any measure function in BayPOD-AL. More specifically, due to the availability of parameter posterior distributions in BayPOD, there is flexibility in defining uncertainty-based measure functions. For BayPOD-UAL, %assuming the optimality of the model leads to a near zero bias, 
we define a measure function dependent on each snapshot's predictive posterior variance. Specifically, using Monte Carlo~(MC) sampling we estimate $\hat{V}^{(t)}(\mathbf{p_U}, x)$, the sample variance of each input $\mathbf{p_U}$ at the specific time point $t$ and position $x$. Finally, by setting 
% the measure function 
$M=\hat{V}$ %and plugging 
in~\eqref{bpod-al}, $\mathbf{p}^*$ can be found by~\eqref{p-star}.

\section{BayPOD-EAL: Error-guided AL}\label{EAL}
Due to the potential mismatch between ROMs and their corresponding high-fidelity model solution, BayPOD-UAL may lead to degraded AL performance as shown in other ML problems \citep{munjal2022robustreproducibleactivelearning, unknown, rahmati2024understandinguncertaintybasedactivelearning}. Consequently, %in the following, 
we develop BayPOD-EAL, which benefits from an error-dependent measure function. 
%, is to find its 
In this approach, to define the measure function we aim to %need to 
estimate the ROM approximation error by its bounds. %One way is to estimate the error bounds. 
% A recent study 
\citet{b713455c1b2c4ae28448b77823fe2a43} approximates the error upper bounds {following the assumption that the underlying ground-truth function can be modeled based on a Gaussian Process (GP).}
%by modeling the underlying governing function with a Gaussian Process~(GP). %Taking advantage of their results,  provide 
We define a similar error-guided measure function and name the corresponding approach BayPOD-EAL. 
Considering $\Tilde{f}^{(t)}_{\mathbf{p_U}x}$ as the model's prediction for unlabeled input $\mathbf{p_U}$, at time $t$ and position $x$, the primary objective is to estimate $(f_{\mathbf{p_U}x}^{(t)}-\Tilde{f}_{\mathbf{p_U}x}^{(t)})^2$, where $f_{\mathbf{p_U}x}^{(t)}$ is the ground truth or FOM solutions. To estimate the error upper bound, \citet{b713455c1b2c4ae28448b77823fe2a43} first 
%model the governing function 
{considers that the ground-truth function} $f$ {can be written} as a GP with a symmetric, positive definite kernel. {Based on this reformulation of $f$}, conditioned on $\mathbf{D_L}$, the posterior distribution, $\mathcal{F}^\prime$, over $f$, 
%is derived and then 
{can be} utilized to find $U^{(t)}(\mathbf{p_U},x)$, the upper-bound of the expected posterior loss $L^{(t)}(\mathbf{p_U},x) = E_{f\sim \mathcal{F}^\prime}[(f_{\mathbf{p_U}x}^{(t)}-\Tilde{f}_{\mathbf{p_U}x}^{(t)})^2]$ by further assuming that kernels are continuously twice differentiable and translation invariant and the governing function's variance is bounded. Utilizing the model's posterior, $q(\cdot|\mathbf{D_L})$, and by representing the model's prediction with its estimated mean via MC sampling, we set %the measure function 
$M=U$ in \eqref{bpod-al} which is then used in \eqref{p-star} to find $\mathbf{p}^*$.

\section{Rod Temperature Evolution}\label{heated rod}
Depending on the heat diffusivity parameter, $\kappa$, the heat diffusion over a rod is governed by: 
\begin{equation}
\frac{\partial f}{\partial t} = \kappa \frac{\partial^2f}{\partial x^2},
\end{equation}
for which the initial condition and Dirichlet boundary conditions in our experiments are set as $f(x=0, t)=3\sin(2t), f(x=L,t)=3, \text{ and } f(x,t=0)=0$, respectively. {These constraints, are incorporated through the particular solution, $\Bar{\boldsymbol{f}}$. $\Bar{\boldsymbol{f}}$ can be derived by solving the problem with boundary conditions $f(x=0, t)=0$ and $f(x=L, t)=1$, and the problem with the boundary condition $f(x=0, t)=1$ and $f(x=L, t)=0$, which we denote as the steady-state solution $\Bar{\boldsymbol{f}}_L(x)$ and $\Bar{\boldsymbol{f}}_0(x)$, respectively. Finally, the corresponding \emph{particular solution} can be written as $\Bar{\boldsymbol{f}} = 3\sin(2t)\Bar{\boldsymbol{f}}_0(x) + 3\Bar{\boldsymbol{f}}_L(x)$~\citep{10105938}.} 
% Deriving the steady-state solutions for two ends of the rod, denoted as $\Bar{\boldsymbol{f}}_0(x)$ and $\Bar{\boldsymbol{f}}_L(x)$, the corresponding \emph{particular solution} for such a setting is $\Bar{\boldsymbol{f}} = 3\sin(2t)\Bar{\boldsymbol{f}}_0(x) + 3\Bar{\boldsymbol{f}}_L(x)$~\citep{10105938}. 
{This allows embedding physics when training} 
BayPOD by learning based on the modified snapshots with homogeneous boundary conditions that are acquired by subtracting snapshots' particular solutions from them. To accurately model dynamics with inhomogeneous boundary conditions, the final approximation is constructed by %~\eqref{eq:approxpod} 
{adding back the \emph{particular solution}}.\vspace{-2mm}

\section{Experiment Settings}\label{exp-settings}\vspace{-2mm}
In all our experiments, each snapshot is a ${n_x}$-dimensional vector, with $n_x=200$, for a temperature field over the rod with diffusivity $\kappa$ at specific time $t$. We consider evaluating BayPOD-AL with 90 equidistant values as diffusivity parameters in $[0.1,0.9]$,  $\mathrm{K}$.
For both AL algorithms, BayPOD-UAL and BayPOD-EAL, we prioritize corresponding diffusivity parameter values in $\mathrm{K}$ based on the experimental settings detailed below. During the AL process, by considering $n_t^{\mathbf{p}}=50$ fixed time points that are randomly chosen from 628 equidistant temporal points in $[0,2\pi]$, which mimics the case that coarse-grained FOM snapshots are used for ROM learning to further improve computational efficiency. At each step of AL, the 50 new snapshots corresponding to {previously chosen fixed time points with} the selected informative diffusivity parameter ($\kappa^*$) are added to $\mathbf{D_L}$. 
%By randomly choosing a parameter from 
The performance comparison experiments are based on random runs starting from a randomly chosen diffusive parameter value in 
$\mathrm{K}$, %the process begins 
with an initial set $\mathbf{D_L}$ that contains 50 snapshots.

%To evaluate performance, w
We focus on an extrapolation setting for performance evaluation with respect to input diffusive parameters, where the snapshots corresponding to the last 20 parameters in $\mathrm{K}$ are chosen as the test dataset. To further investigate the robustness of BayPOD-AL methodologies, we report the performance on two test datasets with different number of time points: 1) $n_t^{\mathbf{p}}=50$ (the same number of time points as in training, 1000 snapshots in total), and 2) $n_t^{\mathbf{p}} = 200$ (4000 snapshots in total) randomly chosen from $T$, for which we refer to as `low' and `high' temporal resolution datasets respectively. 
To account for the objective, the acquisition function~\eqref{bpod-al} is calculated by considering the low-resolution setting ($n_t^\mathbf{p}=50$ per unlabeled input) for experiments on the low-resolution test dataset, and the high-resolution setting ($n_t^\mathbf{p}=200$ per unlabeled input) for experiments on the high-resolution test dataset. 
In this work, the dimension of POD bases is set to 7. 

All the experiments are performed using 10 GB memory on a 40 GB A100 GPU with each experiment taking less than 7 hours. 

%%%%%%%%%%%%%%%%%%%%%%%%%%%%%%%%%%%%%%%%%%%%%%%%%%%%%%%%%%%%

\newpage

\end{document}